\documentclass[letterpaper, 10pt, conference]{ieeeconf}  % Temporary replacement for ieeeconf
\IEEEoverridecommandlockouts
\usepackage{graphicx}
\usepackage{subcaption}
\usepackage{float}
\usepackage{amsmath}
\usepackage{algorithm}
\usepackage{algpseudocode}
\title{\LARGE \bf
Pose State Perception of Interventional Robot for Cardio-cerebrovascular Procedures
}

\author{Shunhan Ji$^{1}$, Yanxi Chen$^{1}$, Zhongyu Yang$^{1}$, Quan Zhang$^{1}$, Xiaohang Nie$^{1}$, Jingqian Sun$^{1\dagger}$, Yichao Tang$^{1\dagger}$ 
\thanks{$^{1}$Tongji University, Shanghai, China}
\thanks{$^{\dagger}$Corresponding author. Email: jingqian\_sun@tongji.edu.cn; tangyichao@tongji.edu.cn}
}

\begin{document}

\maketitle
\thispagestyle{empty}
\pagestyle{empty}

%%%%%%%%%%%%%%%%%%%%%%%%%%%%%%%%%%%%%%%%%%%%%%%%%%%%%%%%%%%%%%%%%%%%%%%%%%%%%%%%

\begin{abstract}

In response to the increasing demand for cardio-cerebrovascular interventional surgeries, precise control of interventional robots has become increasingly important. Within these complex vascular scenarios, the accurate and reliable perception of the pose state for interventional robots is particularly crucial. This paper presents a novel vision-based approach without the need of additional sensors or markers. The core of this paper’s method consists of a three-part framework: firstly, a dual-head multitask U-Net model for simultaneous vessel segment and interventional robot detection; secondly, an advanced algorithm for skeleton extraction and optimization; and finally, a comprehensive pose state perception system based on geometric features is implemented to accurately identify the robot's pose state and provide strategies for subsequent control. The experimental results demonstrate the proposed method's high reliability and accuracy in trajectory tracking and pose state perception. 

\end{abstract}

%%%%%%%%%%%%%%%%%%%%%%%%%%%%%%%%%%%%%%%%%%%%%%%%%%%%%%%%%%%%%%%%%%%%%%%%%%%%%%%%
\section{INTRODUCTION}

The cardiovascular system plays a vital role in supplying oxygenated blood and essential nutrients to the heart and brain. Cardiovascular and cerebrovascular diseases (CCVDs), caused by various risk factors, exhibit high incidence and mortality rates, with cardiovascular diseases being the leading cause of death worldwide, contributing to 31.5\% of all deaths\cite{c1}. While minimally invasive catheter-based procedures have become the preferred treatment over traditional open-heart surgery\cite{c2}, severe critical challenges persist in clinical practice. Primary concerns include radiation exposure risks to medical personnel\cite{c3}, the requirement for high-level operator expertise in catheter manipulation, and persistent dependence on additional sensing equipment even in systems with autonomous capabilities.

Robot-assisted technology has emerged as a promising solution, enabling medical personnel to operate in radiation-free environments. Commercial systems such as the Sensei X robot\cite{c4} and the AMIGO system\cite{c5} have successfully demonstrated remote operation capabilities, though they still require constant manual control without achieving autonomous operation. In the academic domain, progress has been made toward automation, as evidenced by innovative systems including a remote-controller robotic system by Hao shen et al. \cite{c24}, an cardiovascular interventional surgery robot by Wang Shuang et al. \cite{c25}, and a cloud communication-based interventional robot system by Yang Cheng et al. \cite{c26}. However, these autonomous solutions typically rely on additional sensing equipment, which increases system complexity and potentially limits their clinical adoption.

This study proposes a novel pose state perception approach that could potentially improve the level of automation and intelligence of interventional surgical robots. Composed to previous methods that rely on external markers\cite{c22} or sensors\cite{c13, c16}, this approach only utilizes images, thus improving clinical adaptability without introducing any additional system complexity. By integrating vascular segmentation and interventional robot detection into a unified dual-head multitask framework, the method achieves an optimal balance between computational efficiency and accuracy. The study focuses on medical vascular image processing, vascular skeleton extraction, and interventional robot pose state identification, demonstrating superior performance in handling complex vascular structures and real-time processing demands. This advancement could enable an interventional robot to self-adjust their spatial states based on real-time visual feedback, incorporating angle and center line distance calculations to prevent vascular wall collisions and enhance surgical safety. Such innovations lay a solid foundation for closed-loop control, while reducing hardware dependency and improving cost-effectiveness, potentially catalyzing the development of more autonomous interventional systems.

%%%%%%%%%%%%%%%%%%%%%%%%%%%%%%%%%%%%%%%%%%%%%%%%%%%%%%%%%%%%%%%%%%%%%%%%%%%%%%%%

\section{RELATED WORKS}
\subsection{Vassal Segmentation}
In the realm of cardiovascular and cerebrovascular interventions, accurate segmentation of blood vessels are crucial for robot-assisted procedures. Traditional methods, such as Aylward et al. \cite{c100}, which employ dynamic scale adjustment for ridge traversal, and the work of Bashir et al. \cite{c101}, which process a active contour model, laid solid foundations for the field. But they require costly scale-space computations and are not well-suited for complex vascular structures. More recently, medical segmentation has been revolutionized by deep learning methods, such as U-Net \cite{c6}, Med3D \cite{c61}, V-Net \cite{c62} , and a series of DeepLab \cite{c63}. Among these, U-Net and its variants, including Attention U-Net \cite{c7}, Res-U-Net \cite{c8}, UIU-Net \cite{c81} and MSR U-Net \cite{c9}, have become mainstream methods for vascular segmentation due to their remarkable accuracy in complex vascular structures. While achieving impressive accuracy, they exhibit time complexity. Therefore, one of the major challenges in using segmentation results to assist in surgical procedures is to find the balance between high-precision segmentation and real-time processing. 

%Skeletonization methods, such as the traditional approach of Aylward et al. \cite{c100}, which uses dynamic scale adjustment for ridge traversal, and the work of Guedri et al. \cite{c110}, which applies a mathematical approach to track the centerline of retinal blood vessels using skeletonization and pixel classification, have been key advancements. But these methods require costly scale-space computations, while morphological thinning offers computational efficiency (processing 512×512 images in 10ms \cite{van2014scikit}) at the cost of sensitivity to boundary noise and bifurcation artifacts. 

%To address these issues, this study proposes a dual-head multitask U-Net framework and uses graph-based pruning functions to eliminate bifurcation artifacts from the skeleton generated by morphological thinning, balancing accuracy and efficiency,从而为实时微创介入手术提供了更可靠的支持。

\subsection{Pose State Perception of the Robot} 
In minimally invasive interventional surgery, the precise pose state perception of the interventional robot is crucial for surgical safety. Existing methods can be classified into three main types: (a) image-based visual methods \cite{c10, c15}, (b) electromagnetic track \cite{c16,c11,c23}, and (c) sensing of actuation fibers \cite{c13, c131}. Although the latter two methods ensure a certain level of perception accuracy, they are obviously limited by environmental or device constraints. The second method obtains the interventional robot's pose information through electromagnetic induction technology, but it is easily affected by external magnetic materials in the environment. The third method captures continuous stress changes through actuation fibers to calculate the deformation of the robot \cite{c132, c133}. However, it is hard to integrate actuation fibers with submilimeter-scale interventional devices due to the size limitation of these sensors. Moreover, both of these two methods can only detect the robot's absolute position and cannot directly determine its relative pose within the vessel. Additional registration between the vessel's and sensor's position further increases positioning errors. In contrast, the first method which combines deep learning with visual tracking, by eliminating the need for additional sensors, enables a simpler and more compact robot design, which in turn facilitates more precise and delicate surgeries \cite{c17,c18,c20}.
%1.绝对位置 2.配对误差 3.加深定位误差且上述两种方法只能探测到机器人的绝对位置，无法直接明确其在血管中的相对位姿，需要另外重构血管位置来进行配对定位，由此加深了定位误差。
%This work falls into the image-based category and distinguishes itself from previous image-based methods relying on markers \cite{c22} by solely utilizing images, greatly improving the system's adaptability and widespread deployment potential.

%%%%%%%%%%%%%%%%%%%%%%%%%%%%%%%%%%%%%%%%%%%%%%%%%%%%%%%%%%%%%%%%%%%%%%%%%%%%%%%%
\section{MATERIAL \& METHODS}
In this study, to achieve precise pose state perception of the interventional robot, a comprehensive system is designed and implemented based on image processing. The method consists mainly of three key steps: First, a dual-head multitask U-Net framework is constructed to simultaneously complete vascular segmentation and robot detection. Second, their skeletons are extracted and algorithms are used to filter the robot’s trajectory to obtain the correct one. Finally, a line fitting algorithm is applied to calculate the deviation between the interventional robot’s head and the vascular centerline, adjusting the robot’s movement direction accordingly.

\subsection{Experimental equipment and data sets}
\begin{figure}[htbp]  
\centering  
\includegraphics[width=0.5\textwidth]{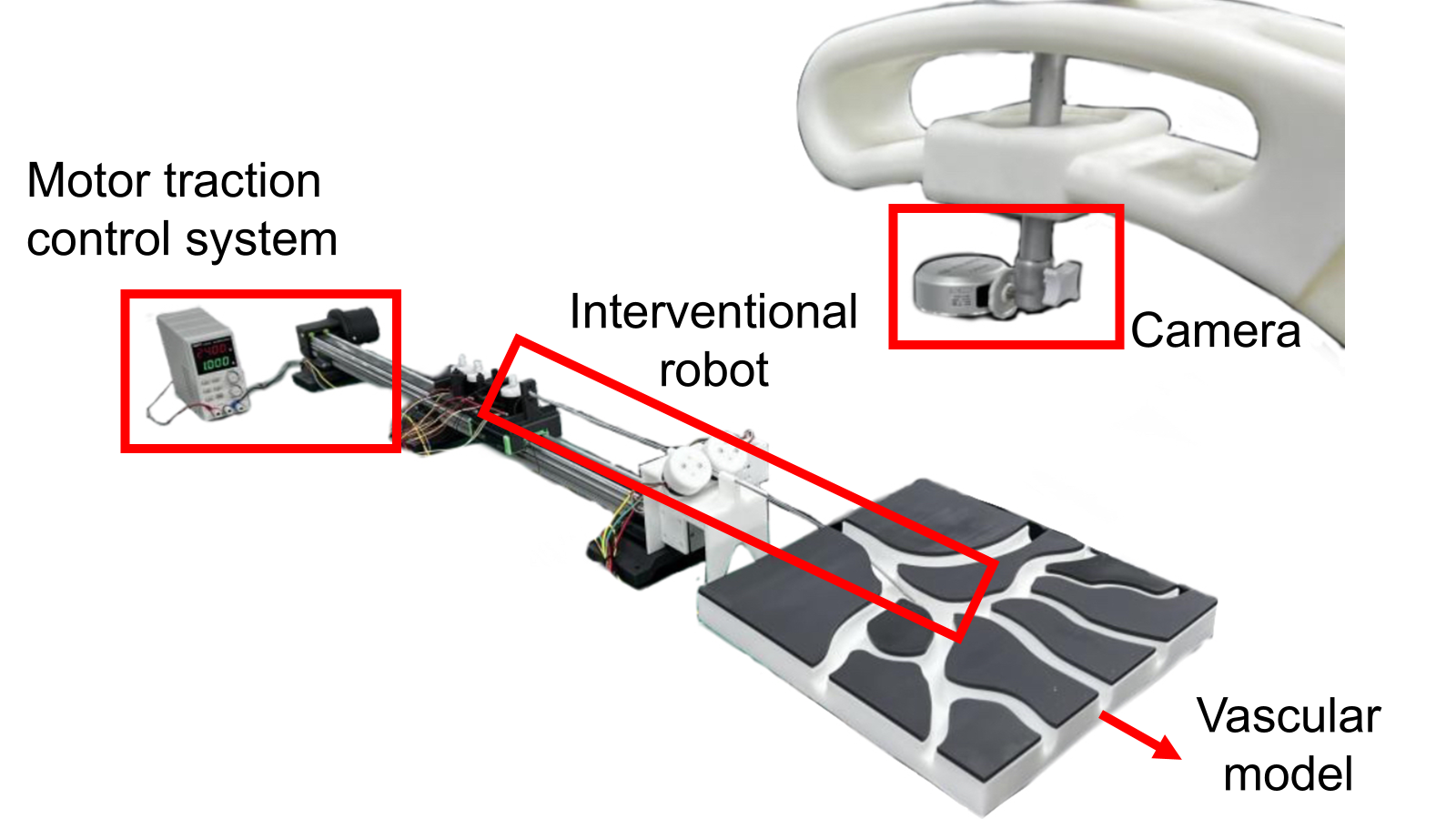}  
\caption{The interventional robot system}  
\label{fig:system}  
\end{figure}

The experimental setup uses an interventional robot system with motor traction control, independently built by the laboratory (Fig. \ref{fig:system}). Four servomotors are used for the four guidewires' traction, precisely controlling the direction of the robot's head. An additional servomotor controls the forward movement of the robot module, while two extra servomotors are used for catheter delivery. The experiment utilizes a vascular model based on real blood vessel design, mimicking the actual operational environment. Image capture is performed using a RealSense L515 RGB camera (Intel Co., Ltd., CA) for real-time image acquisition. The captured images are then manually annotated to provide accurate label information for the dataset. Finally, annotation files are uniformly converted, forming the dataset used in the experiment.

\subsection{Vascular Structure Segmentation and Interventional Robot Detection}
To achieve efficient and precise autonomous control of the interventional robot, preprocessing of the images is required to avoid significant errors caused by noise, low contrast, and other factors when directly processing raw images. This involves segmentation to obtain a binary image, which extracts vessel and catheter.

This section of the study is inspired by the work of Zhihui Guo et al. In a similar vein to the idea of their paper "DeepCenterline: A Multi-task Fully Convolutional Network for Centerline Extraction."\cite{c17}, a dual-head multi-task U-Net framework is employed. The dual-head architecture enables concurrent vascular structures segmentation and interventional robot detection, leveraging shared low-level features to enhance both the accuracy and robustness of the model. It fully exploits the relationships between the two tasks, improving both the segmentation accuracy of the vascular structures and the interventional robot, and enhancing the model’s performance in complex medical images.

\begin{figure}[htbp]  
\includegraphics[width=0.45\textwidth]{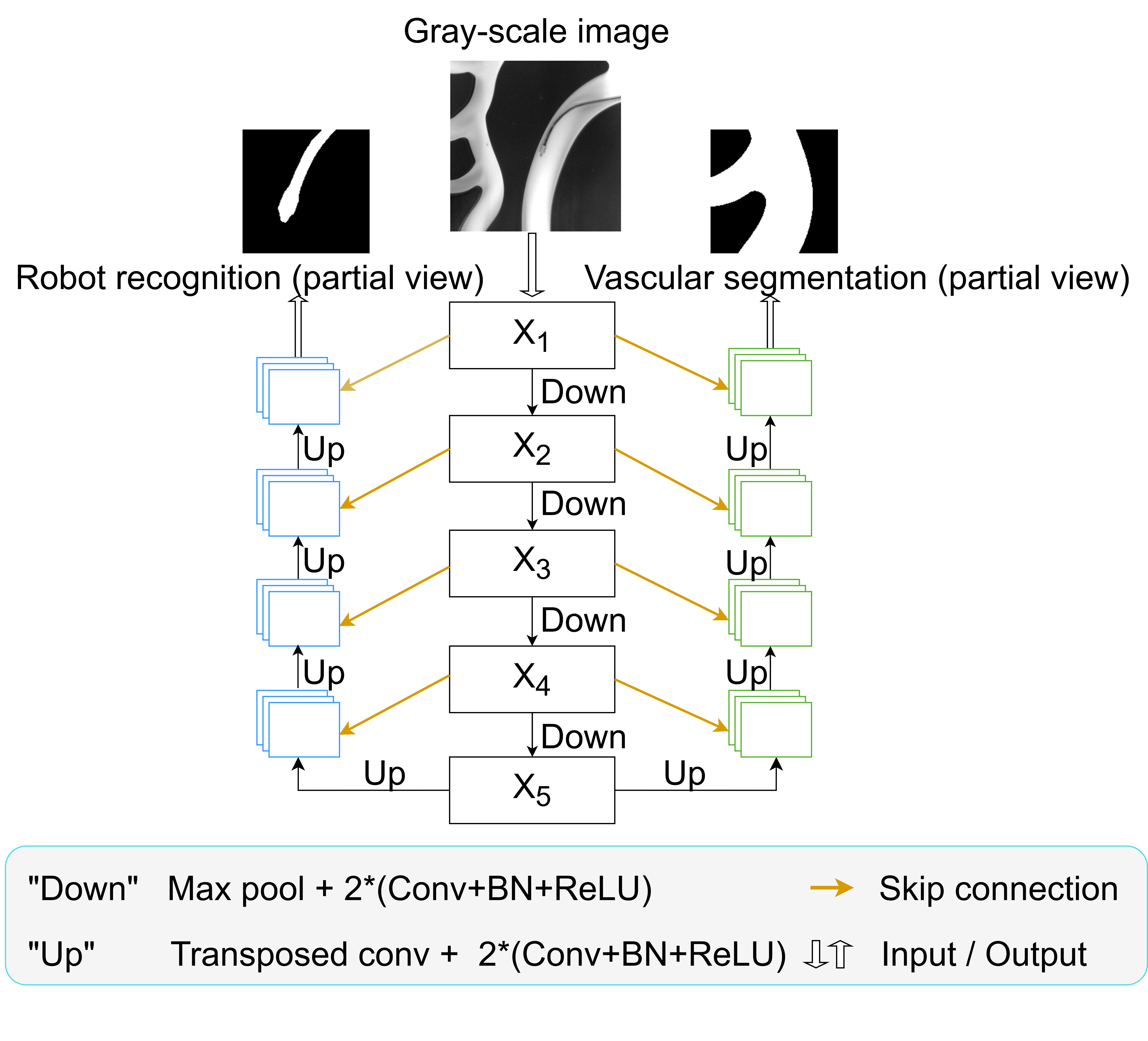} 
\caption{Architecture of the proposed multi-task U-Net. Both tasks share the same downsampling process to abstract the image and extract local features through operations such as convolution and pooling. Subsequently, each task has its own distinct upsampling component, including transposed convolutions, to recover feature information and generate the final output.} 
\label{fig:dual}  
\end{figure}

The U-Net architecture in this study comprises four downsampling layers and four upsampling layers (Fig. \ref{fig:dual}). The downsampling path incorporates max-pooling operations and dual convolution blocks, each with batch normalization and ReLU activation, to extract hierarchical features while reducing computational complexity. The upsampling path utilizes transposed convolutions with skip connections to restore spatial resolution while preserving fine-grained details from earlier layers, effectively addressing the vanishing gradient problem.

\begin{figure}[htbp]
\centering
\begin{subfigure}[b]{0.15\textwidth}  
  \includegraphics[width=\textwidth]{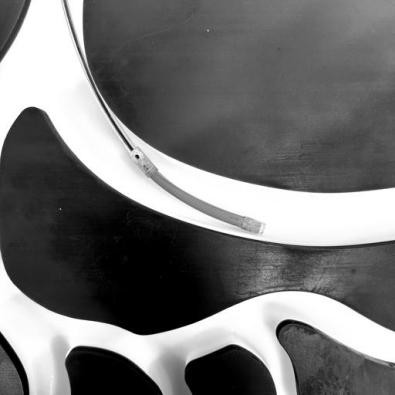}  
  \caption{}
  \label{fig:subfig1}
\end{subfigure}
\hfill
\begin{subfigure}[b]{0.15\textwidth}
  \includegraphics[width=\textwidth]{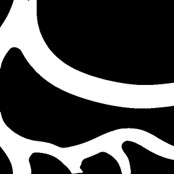}
  \caption{}
  \label{fig:subfig2}
\end{subfigure}
\hfill
\begin{subfigure}[b]{0.15\textwidth}
  \includegraphics[width=\textwidth]{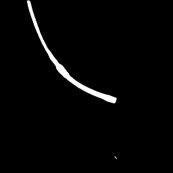}
  \caption{}
  \label{fig:subfig3}
\end{subfigure}
\caption{Accurate Segmentation of Vascular Region and interventional robot Using Dual-Head Multi-Task U-Net Framework. (a) The original image; (b)Binary Image After Vascular Segmentation; (c)Mask Image of interventional robot Detection}
\label{fig:three_images}
\end{figure}

With this dual-head multitask framework, the raw image is converted into a binary segmented image of the blood vessels and the mask image for interventional robot detection (Fig. \ref{fig:three_images}), providing a reliable foundation for the subsequent image processing steps.

\subsection{Vessel and interventional robot Skeleton Extraction and Track Filtering}
To accurately extract the skeleton of the interventional robot and ensure the continuity of its path, the segmented binary vascular image is first preprocessed to remove any potential small noise regions, which are typically caused by incomplete segmentation or local image noise. For this purpose, functions are used to eliminate small connected areas and holes, resulting in a cleaner vascular shape. 

\begin{figure}[htbp]
\centering
\begin{subfigure}[b]{0.22\textwidth}  
  \includegraphics[width=\textwidth]{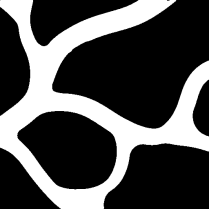} 
  \caption{A segmented vessel image}
  \label{fig:subfig1}
\end{subfigure}
\hfill
\begin{subfigure}[b]{0.22\textwidth}
  \includegraphics[width=\textwidth]{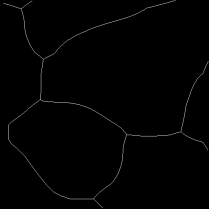}
  \caption{Extracted skeleton of (a)}
  \label{fig:subfig2}
\end{subfigure}
\caption{Result of skeletonize}
\label{fig:skeletonize}
\end{figure}

After removing the noise, skeletonization function is applied to process the image. This method is based on morphological operations, gradually eroding the image to extract a 1-pixel wide centerline while preserving topological connectivity, which ensures good continuity and maintains high computational efficiency. The skeletonized image retains the primary structure of the vessel and the robot, removing extraneous pixels to form a simplified yet representative skeleton (Fig. \ref{fig:skeletonize}), which facilitates subsequent analysis. 

However, due to image noise or resolution limitations, local fractures may still occur during the skeletonization process. To address the problem, a repair algorithm based on the local structural features of the skeleton has been designed, which detects the end points in the skeleton using convolution operations and automatically connects the broken end points whose distance is less than a predefined threshold, thereby restoring the skeleton’s integrity. Algorithm \ref{algorithm:connect} shows the pseudocode:

\begin{algorithm}
\caption{Connect Skeleton Gaps}
\begin{algorithmic}[1]
\small
\State \textbf{Input:} skeleton: binary skeleton image, gap\_threshold: maximum distance to connect gaps
\State \textbf{Output:} processed skeleton image with gaps connected
\State Initialize kernel $K$ as $3 \times 3$ matrix: 
\[
K = \begin{bmatrix} 
1 & 1 & 1 \\
1 & 10 & 1 \\
1 & 1 & 1 
\end{bmatrix}
\]
\State Calculate $conv = \text{convolution on skeleton with kernel} K$
\State Calculate endpoints = positions where $(conv = 11) \text{ AND } (skeleton = 1)$
\State Convert endpoints to $(x, y)$ coordinate list
\For{each endpoint $i$ in endpoints}
    \State $p_1 = \text{endpoints}[i]$
    \For{each endpoint $j$ in endpoints starting from $i$}
        \State $p_2 = \text{endpoints}[j]$
        \State Calculate $dist = \text{Euclidean distance between } p_1$
        \State \quad $\text{ and } p_2$
        \If{$dist < gap\_threshold$}
            \State Connect $p_1$ and $p_2$ on result image
        \EndIf
    \EndFor
\EndFor
\end{algorithmic}
\label{algorithm:connect}
\end{algorithm}

\begin{figure}[htbp]
\centering
\includegraphics[width=0.45\textwidth]{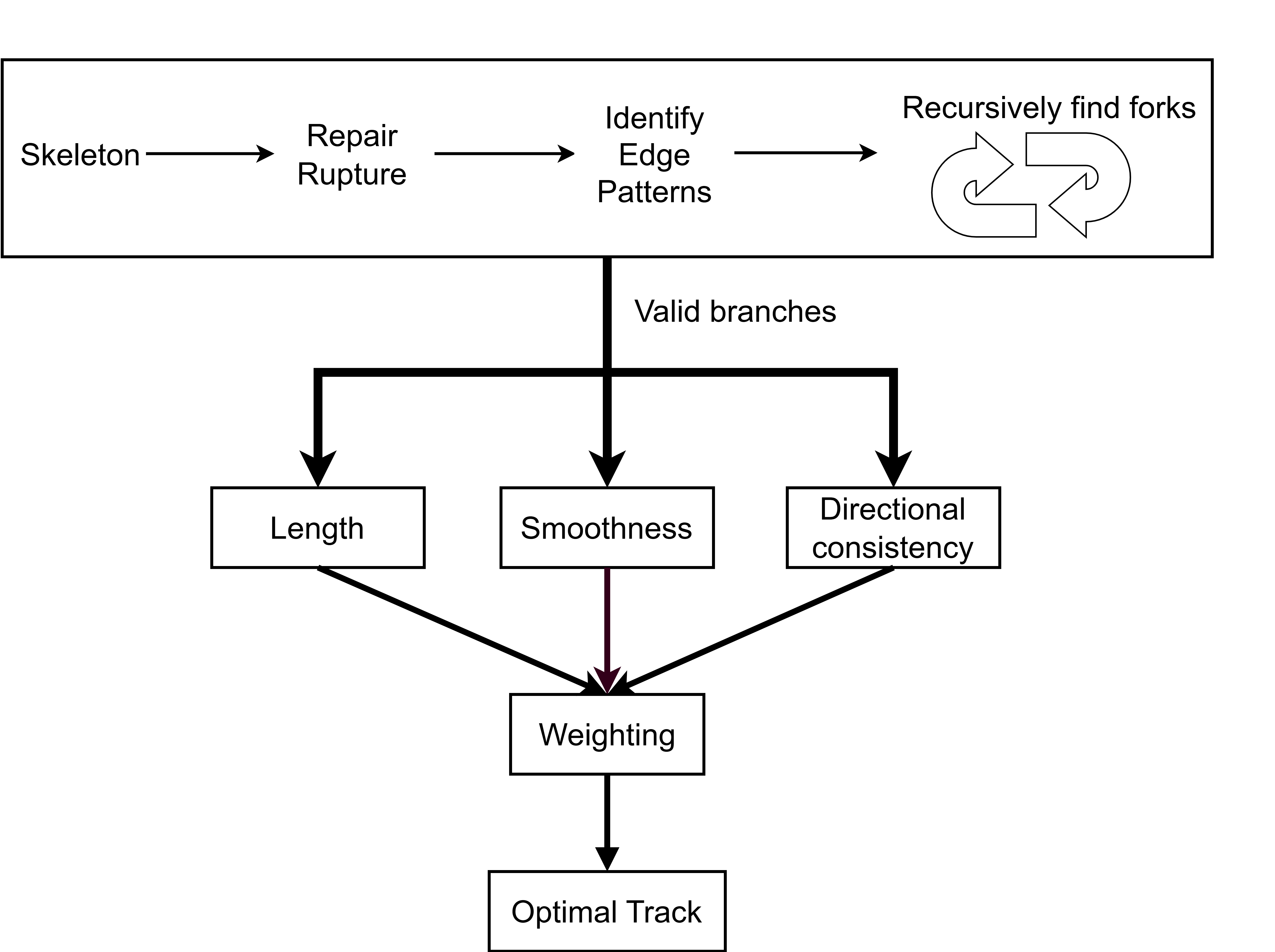} 
\caption{Determine the optimal trajectory of interventional robot}
\label{fig:track_deter} 
\end{figure}

In alignment with the precision required in the study, it is essential to accurately extract the skeleton of the interventional robot, ensuring the continuity of the path and eliminating any potential branches or fractures, while precisely locating the robot’s head. To achieve this goal, a multi-stage path filtering algorithm has been designed. By setting appropriate thresholds, the algorithm recursively processes branching paths and comprehensively considers multiple feature dimensions of the path, ultimately determining the unique and optimal skeleton path for the robot. The flow is shown in Fig. \ref{fig:track_deter}.

First, Algorithm \ref{algorithm:connect} repairs the skeleton fractures to ensure path continuity. Starting from detected edge points, a recursive path-tracing algorithm is employed to handle any possible branching in the skeleton. The algorithm explores all possible branches while preventing revisitation through a visited-points tracking mechanism. The longest valid branch is ultimately selected as the main path with a maximum recursion depth to avoid loop formation.

Once all possible paths from the starting points are filtered, evaluate three key dimensions to ensure the optimal path is selected:

\begin{itemize}
  \item \textbf{Path Length}: reflects the movement distance of the robot within the vascular system. Shorter paths are generally preferred as they are more likely to represent the actual trajectory of the robot.

  \[
  score_l = \text{length(path)}
  \]
  
  \item \textbf{Smoothness}: measures the degree of curvature variation of the path. In actual motion, the robot tends to move smoothly and continuously, without sharp changes in curvature.

  \[
  score_s = \frac{1}{n-2} \sum_{i=1}^{n-2} \left|\cos(\theta_i) \right|
  \]
  where \(n\) is the number of path points, and \(\theta_i\) is the angle formed between the \(i\)-th point and its adjacent two points.

  \item \textbf{Direction Consistency}: considers the stability of the overall direction of the path. An ideal trajectory should maintain a consistent direction. This metric selects the relatively optimal path by calculating the consistency between each segment’s direction and the overall direction.

  \[
  score_c = \frac{1}{n-1} \sum_{i=1}^{n-1} \left| \cos(\phi_i) \right|
  \]
  where \(\phi_i\) is the angle between the local direction vector \(v_i\) and the main direction vector \(v_{\text{main}}\).
  \[
  v_\text{main}=(x_\text{end}-x_\text{start}, y_\text{end}-y_\text{start})
  \]
  \[
  v_i=(x_\text{i+1}-x_i, y_\text{i+1}-y_i)
  \]
\end{itemize}

\begin{figure}[H]
\centering
\begin{subfigure}[b]{0.15\textwidth}  
  \includegraphics[width=\textwidth]{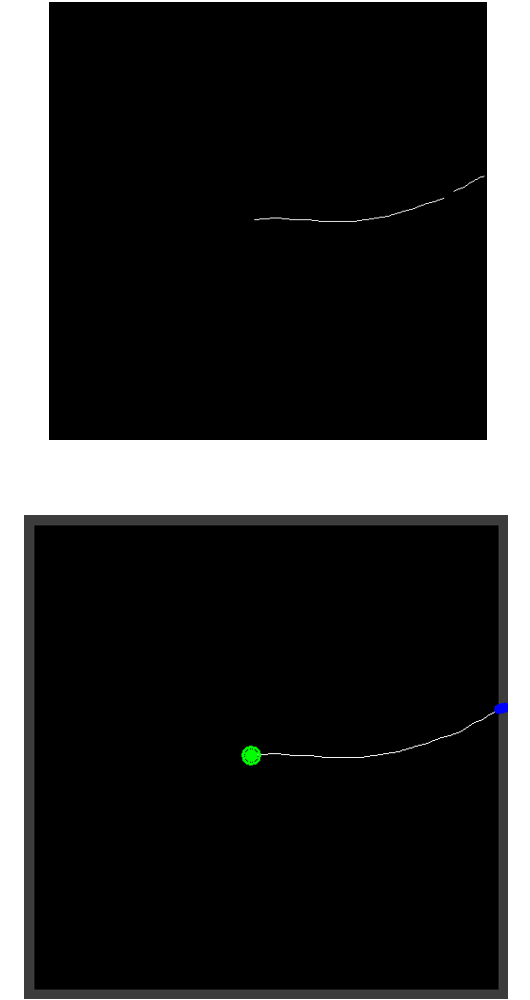} 
  \caption{}
\end{subfigure}
\hfill
\begin{subfigure}[b]{0.15\textwidth}
  \includegraphics[width=\textwidth]{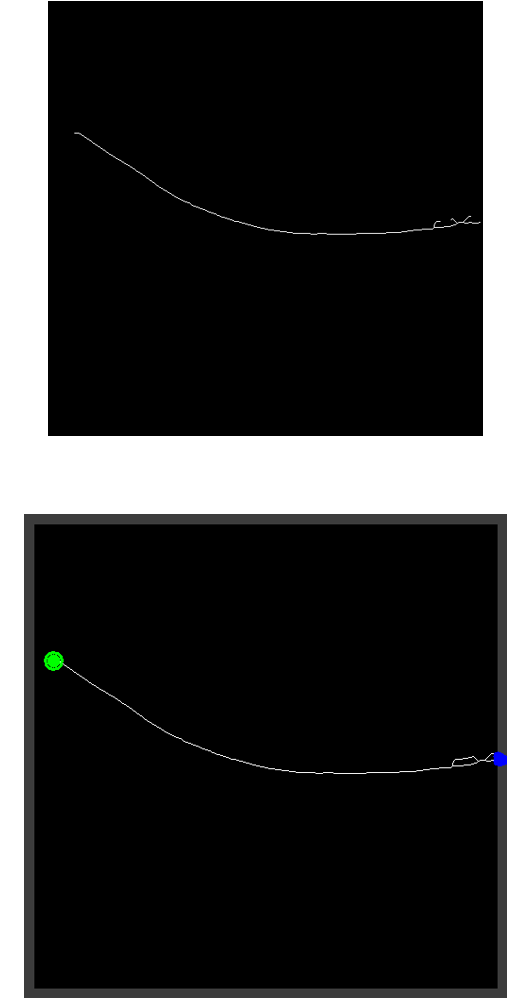}
  \caption{}
\end{subfigure}
\hfill
\begin{subfigure}[b]{0.15\textwidth}
  \includegraphics[width=\textwidth]{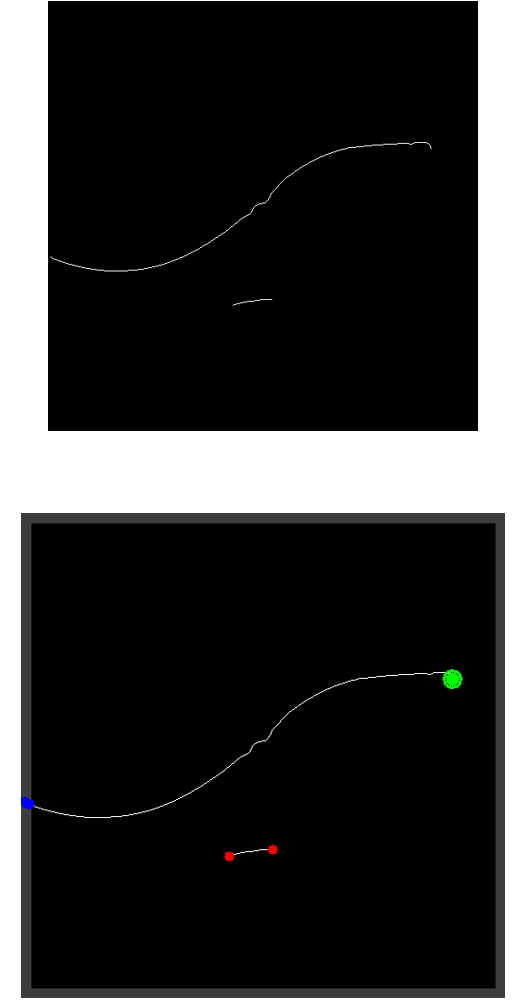}
  \caption{}
\end{subfigure}
\caption{Different situations that algorithm can handle. (a)intermediate break; (b)branches; (c)outlier segments; Green dot represents robot's head, blue dot represents boundary points and red dot represents other end points}
\label{fig:head}
\end{figure}

By scoring from these three dimensions and applying a weighted approach, the algorithm ultimately obtains the optimal path as the true trajectory of the robot. The head coordinates of it are then extracted (Fig. \ref{fig:head}), providing reliable technical support for subsequent pose state estimation.

\[
score=\sum_{i=l, s, c} w_i \cdot score_i
\]

\subsection{Centerline Fitting of Robot's Head and vessel and Pose State Adjustment}
\begin{figure}[htbp]
\centering
\includegraphics[width=0.25\textwidth]{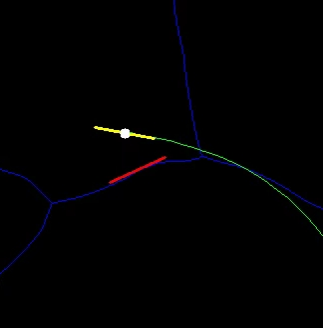} 
\caption{Image of fitted lines}
\label{fig:fitted} 
\end{figure}

Now all the foundational data required for the interventional robot's pose state perception have been obtained. Starting from the detected robot's head point, and trace back 40 pixels for linear fitting, the robot's direction vector is obtained. By overlaying vascular and robot skeletons, the algorithm locates the point on the vascular skeleton nearest to the robot's head, then a 40-pixel (±20 pixels) segment of the skeleton around this point is extracted and fitted to obtain the vessel's direction vector (Fig. \ref{fig:fitted}). Both vectors are represented by their start, center, and end points.

\begin{figure}[htbp]
\centering
\includegraphics[width=0.45\textwidth]{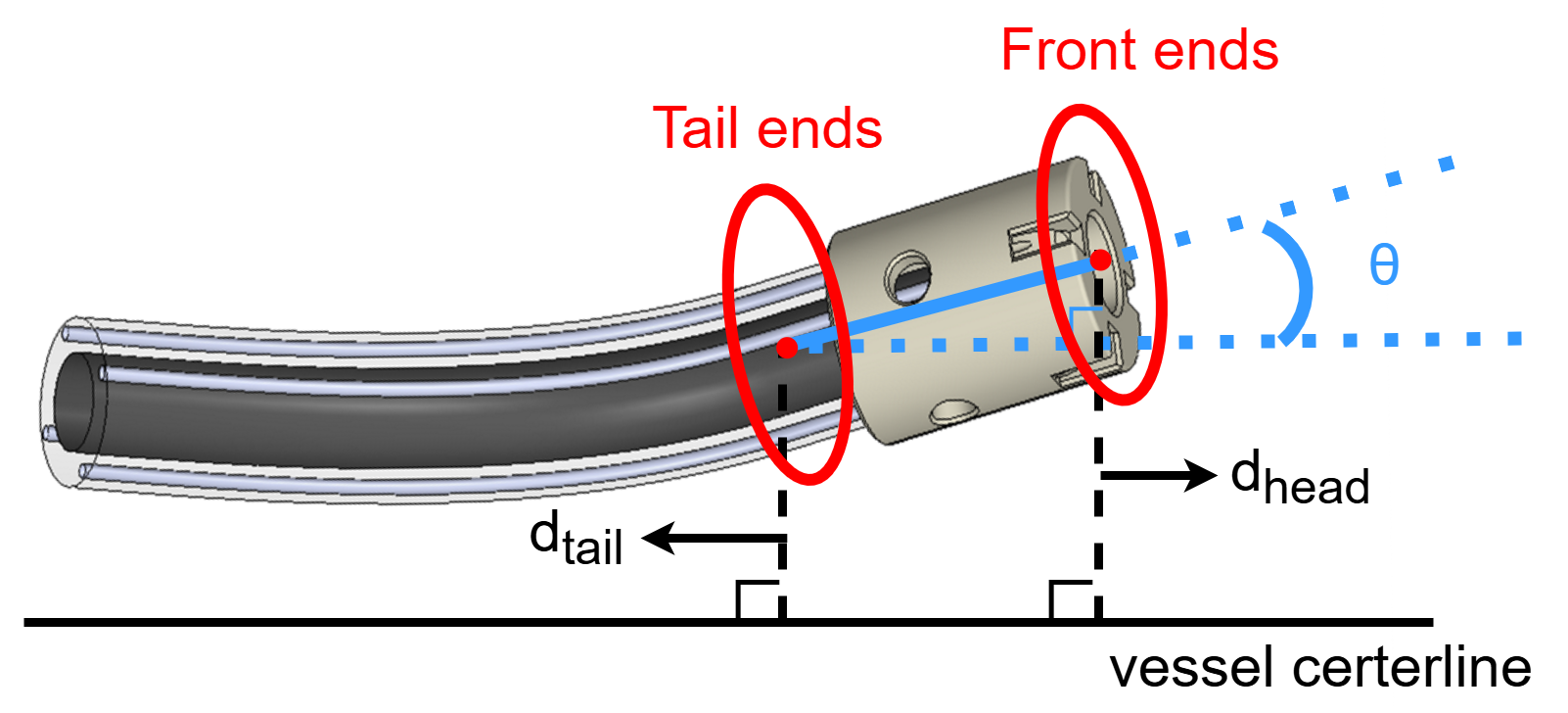} 
\caption{Key parameters of interventional robot}
\label{fig:catheter_para} 
\end{figure}

The robot's pose perception is then conducted from two aspects: relative position, and quantifying the posture through angles. The key parameters (Fig. \ref{fig:catheter_para}) include:\quad a) cross products between vascular vector and robot's front/tail ends ($c_\text{head}$/$c_\text{tail}$);\quad b) absolute values of cross products, representing distances to the vascular centerline ($d_\text{head}$/$d_\text{tail}$);\quad c) XOR of cross product signs, indicating whether centerline crossing ($s$);\quad d) angle between robot and vessel direction vectors ($\theta$).

\begin{figure}[htbp]
\centering 
\includegraphics[width=0.45\textwidth]{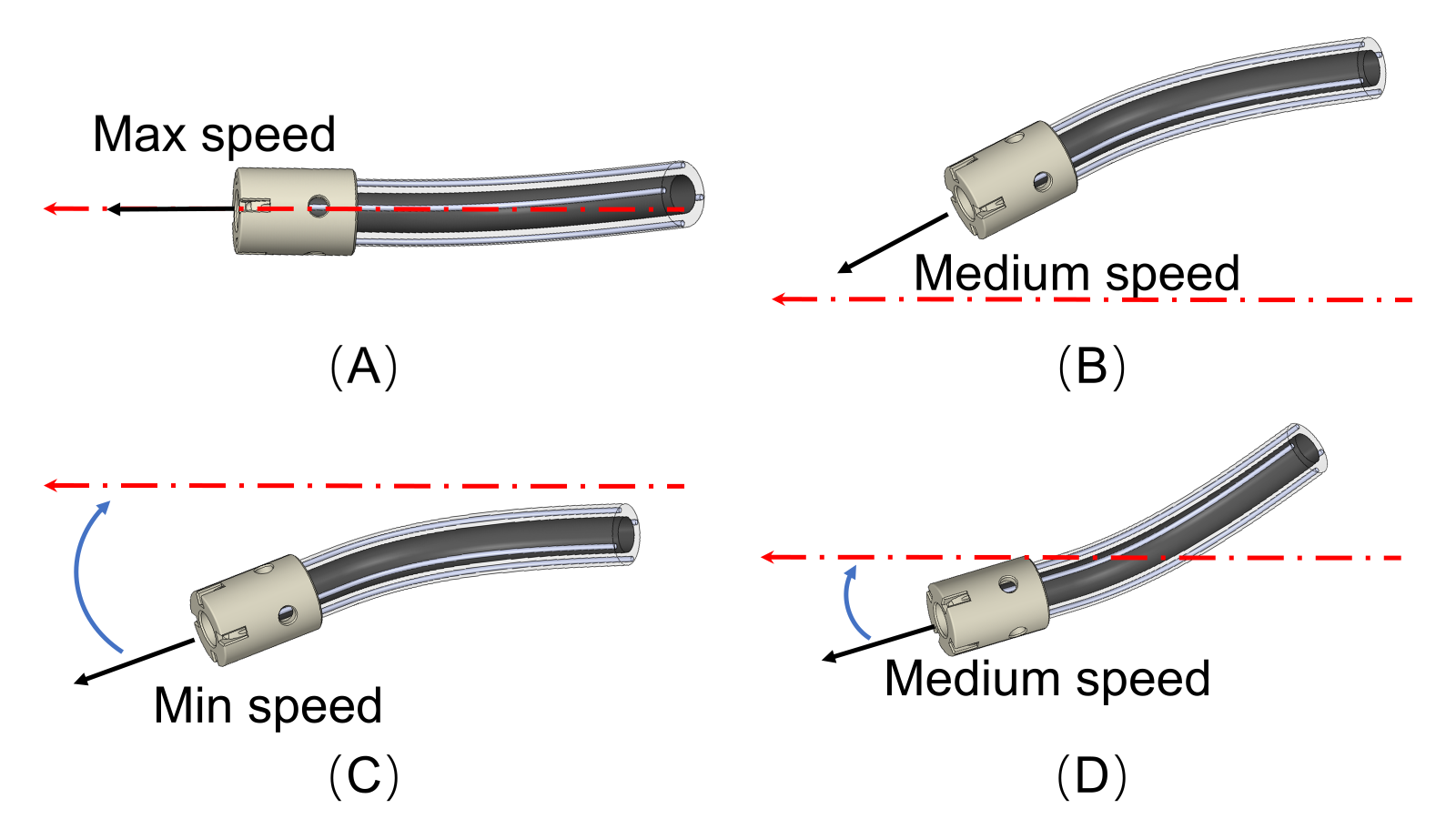} 
\caption{4 states of interventional robot}
\label{fig:4_state} 
\end{figure}

Considering these four parameters - $d_\text{head}$, $d_\text{tail}$, $s$, and $\theta$ - and adhering to the principle of maximizing the robot's forward speed, the robot's pose state is classified into the following situations:

\begin{itemize}
\item When both $d_\text{head}$ and $d_\text{tail}$ are smaller than $d_{\text{allow}}$(preset distance threshold), and $\theta$ is smaller than $\theta_{\text{allow}}$ (preset angle threshold), the state is classified as A (Fig. \ref{fig:4_state}(A)). This indicates that the robot's trajectory is ideal, the speed can be increased, and no steering is needed;

\item When either distance exceeds $d_{\text{allow}}$, $s$ is true, and $d_\text{head}$ is smaller than $d_\text{tail}$, the state is classified as B (Fig. \ref{fig:4_state}(B)). In this case, no direction change is required, but the speed should be reduced for better control;

\item When $s$ is true, but $d_\text{head}$ is greater than $d_\text{tail}$, the state is classified as C (Fig. \ref{fig:4_state}(C)). This suggests that the robot has deviated significantly, and the speed should be reduced to the minimum to allow for maximum steering;

\item When $s$ is false, the state is classified as D (Fig. \ref{fig:4_state}(D)), indicating that the robot's head has crossed the vascular centerline but still has a certain angular deviation. The forward speed should be adjusted to a moderate value, with simultaneous steering.
\end{itemize}

For the last two states requiring steering, the algorithm further incorporates the parameter $\theta$ to control the degree of steering, thereby achieving complete pose state perception of the interventional robot.

%%%%%%%%%%%%%%%%%%%%%%%%%%%%%%%%%%%%%%%%%%%%%%%%%%%%%%%%%%%%%%%%%%%%%%%%%%%%%%%%
\section{RESULTS}
\subsection{Angle Perception}
The algorithm's angle perception accuracy is evaluated by comparing automated calculations with manual measurements. Manual measurement involves annotating the centerlines of both the robot's head and nearby vascular area and calculating the angle. This comparison provides an objective assessment of the algorithm's accuracy.

The study tested 48 representative images, each measured twice manually to minimize the influence of random errors. Through in-depth analysis results, a series of compelling statistical indicators are obtained: mean error of 2.25 degrees, standard deviation of 5.68 degrees, and median error of 2.14 degrees. Considering the full measurement range of 180 degrees, the mean error accounts for only 1.25\% of the total range, demonstrating the algorithm’s considerable measurement precision. These statistics demonstrate a strong alignment between algorithmic and manual measurements, validating the algorithm's angle perception accuracy.

\begin{figure}[htbp]  
\centering  
\includegraphics[width=0.4\textwidth]{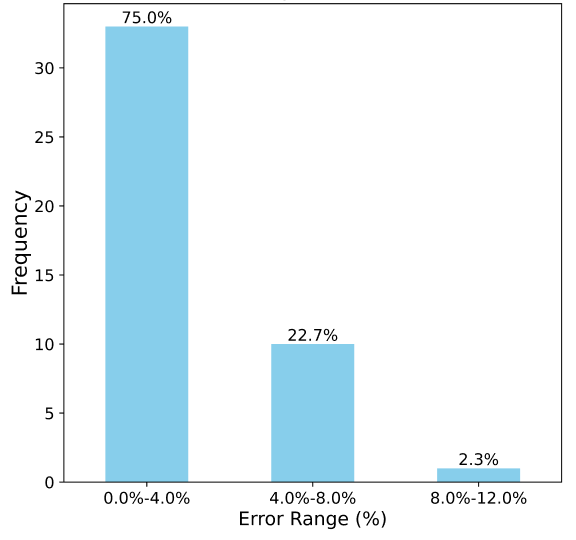} 
\caption{Distribution of error ranges}  
\label{fig:distribution}  
\end{figure}

To more comprehensively evaluate the algorithm’s performance, various statistical methods are applied for analysis. By calculating the Pearson correlation coefficient (\(\tau\)) and the Spearman rank correlation coefficient (\(\rho\)):

\[
\tau = \frac{\sum_{i} \left[ (x_i - \bar{x})(y_i - \bar{y}) \right]}{\sqrt{\sum_{i} (x_i - \bar{x})^2 \cdot \sum_{i} (y_i - \bar{y})^2}}
\]

\[
\rho = 1 - \frac{6 \sum_{i} d_i^2}{n(n^2 - 1)}
\]
In which $d_i$ represents the rank difference and n is the sample size. The analysis yields the two coefficients of 0.869 and 0.870, indicating a strong positive correlation between algorithmic and manual measurements. Furthermore, with percentage-based statistical analysis of the measurement angle errors (Fig. \ref{fig:distribution}), it is discovered that the measurement results demonstrate excellent precision. 75\% of the measurement errors fall within the range of 0-4\%, with an additional 22.7\% of errors within the 4-8\% range.

\begin{figure}[htbp] 
\centering
\includegraphics[width=0.4\textwidth]{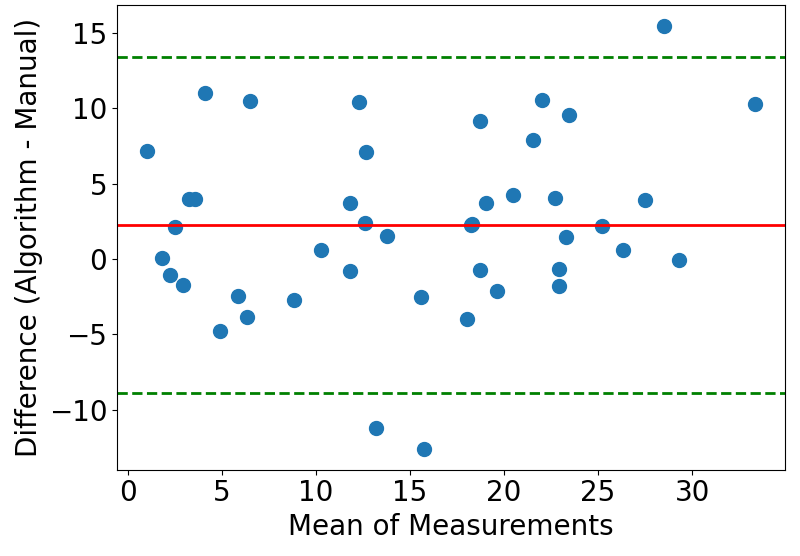}
\caption{Bland-Altman }  
\label{fig:bland}  
\end{figure}

It is particularly noteworthy that in the Bland-Altman analysis (Fig. \ref{fig:bland}), previously calculated mean error (\(\bar{d}\)) and standard deviation of error (\(s\)) are used, and also calculates the 95\% limits of agreement (LoA):

\[
\bar{d} = \frac{\sum_{i} (x_i - y_i)}{n}, \quad s = \sqrt{\frac{\sum_{i} (d_i - \bar{d})^2}{n - 1}}
\]

\[
\text{Upper LoA} = \bar{d} + 1.96s, \quad \text{Lower LoA} = \bar{d} - 1.96s
\]

The figure shows the mean error(red line) and 95\% LoA(green dashed lines). The distribution of data points reveals strong consistency between algorithmic and manual measurements, with most points within the 95\% confidence interval. The absence of systematic bias or trend indicates that the algorithm maintains stable and reliable performance across different angle ranges. Taken together, all evaluation metrics provide strong evidence that the proposed algorithm can serve as a reliable automated measurement method, meeting the requirements for practical applications.

\subsection{Pose State Classification}

To further evaluate the algorithm’s capability to comprehensively perceive the interventional robot’s pose state, the study tests the four pose state classifications mentioned in the previous method section. During the experiment, the algorithm classifies different images, while simultaneously performing manual classification on the same images to validate whether the algorithm’s results align with the expected outcomes.

The evaluation metrics listed in Table \ref{tab:evaluation_metrics} are used in the study:

\renewcommand{\arraystretch}{2}

\begin{table}[ht]
\centering
\begin{tabular}{|l|l|l|}
\hline
\textbf{Metric} & \textbf{Formula} \\ \hline
\textbf{Classification Accuracy (ACC)} & \(\mathrm{ACC} = \frac{\mathrm{TP} + \mathrm{TN}}{\mathrm{TP} + \mathrm{TN} + \mathrm{FP} + \mathrm{FN}}\) \\ \hline
\textbf{Precision (P)} & \(\mathrm{P} = \frac{\mathrm{TP}}{\mathrm{TP} + \mathrm{FP}}\) \\ \hline
\textbf{Recall (R)} & \(\mathrm{R} = \frac{\mathrm{TP}}{\mathrm{TP} + \mathrm{FN}}\) \\ \hline
\textbf{F1 Score} (\( F_1 \)) & \( F_1 = \frac{2 \cdot (\mathrm{P} \cdot \mathrm{R})}{\mathrm{P} + \mathrm{R}} \) \\ \hline
\textbf{Cohen's Kappa (K)} & \(\mathrm{K} = \frac{p_o - p_e}{1 - p_e}\) \\ \hline
\textbf{Fowlkes-Mallows Index (FM)} & \(\mathrm{FM} = \sqrt{\frac{\mathrm{TP}}{\mathrm{TP} + \mathrm{FP}} \cdot \frac{\mathrm{TP}}{\mathrm{TP} + \mathrm{FN}}}\) \\ \hline
\end{tabular}
\caption{Evaluation Metrics and Definitions}
\label{tab:evaluation_metrics}
\end{table}

In this context, the following terms are defined:

\begin{itemize}
  \item \textbf{TP (True Positive)}: The number of samples correctly classified as the target class.
  \item \textbf{TN (True Negative)}: The number of samples correctly classified as the non-target class.
  \item \textbf{FP (False Positive)}: The number of samples incorrectly classified as the target class.
  \item \textbf{FN (False Negative)}: The number of samples incorrectly classified as the non-target class.
  \item \textbf{\( p_o \)}: The observed agreement between the predicted and actual classes.
  \item \textbf{\( p_e \)}: The expected agreement, considering the random chance.
\end{itemize}

\begin{figure}[htbp]
\centering
\begin{subfigure}[b]{0.15\textwidth}  
  \includegraphics[width=\textwidth]{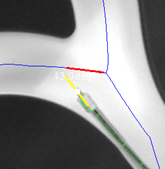}  
  \caption{}
  \label{fig:subfig1}
\end{subfigure}
\hfill
\begin{subfigure}[b]{0.15\textwidth}
  \includegraphics[width=\textwidth]{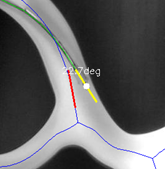}
  \caption{}
  \label{fig:subfig2}
\end{subfigure}
\hfill
\begin{subfigure}[b]{0.15\textwidth}
  \includegraphics[width=\textwidth]{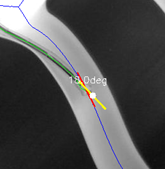}
  \caption{}
  \label{fig:subfig3}
\end{subfigure}
\caption{Automatic classification of B, C, D respectively}
\label{fig:three_images2}
\end{figure}

The results of the experiment show that the algorithm achieves an overall classification accuracy of 88.6\%, along with a Cohen’s Kappa coefficient of 0.796 and an FM index of 0.801. These complementary metrics collectively confirm the strong consistency between the algorithm’s results and manual judgement, with performance significantly surpassing the random classification baseline. It is important to note that Class A(ideal tracks) rarely occurs in actual operations, so the data collected during the experiment does not include any relevant samples. As a result, the final evaluation results only involve the three states: B, C, and D. The figure (Fig. \ref{fig:three_images2}) below provides a clear illustration of the algorithm’s classification performance on these three states. Detailed evaluation metrics are presented in Table \ref{tab:assess_detail}. 

\begin{table}[h!]
\centering
\begin{tabular}{|c|c|c|c|}
\hline
\textbf{Model} & \textbf{Precision} & \textbf{Recall} & \textbf{F1 score} \\ \hline
A & \textbackslash & \textbackslash & \textbackslash \\ \hline
B & 0.86 & 0.92 & 0.89 \\ \hline
C & 0.92 & 0.88 & 0.90 \\ \hline
D & 0.80 & 0.80 & 0.80 \\ \hline
Macro avg & 0.86 & 0.87 & 0.86 \\ \hline
Weighted avg & 0.89 & 0.89 & 0.89 \\ \hline
\end{tabular}
\caption{Assessment details}
\label{tab:assess_detail}
\end{table}

\begin{figure}[htbp]  
\centering  
\includegraphics[width=0.4\textwidth]{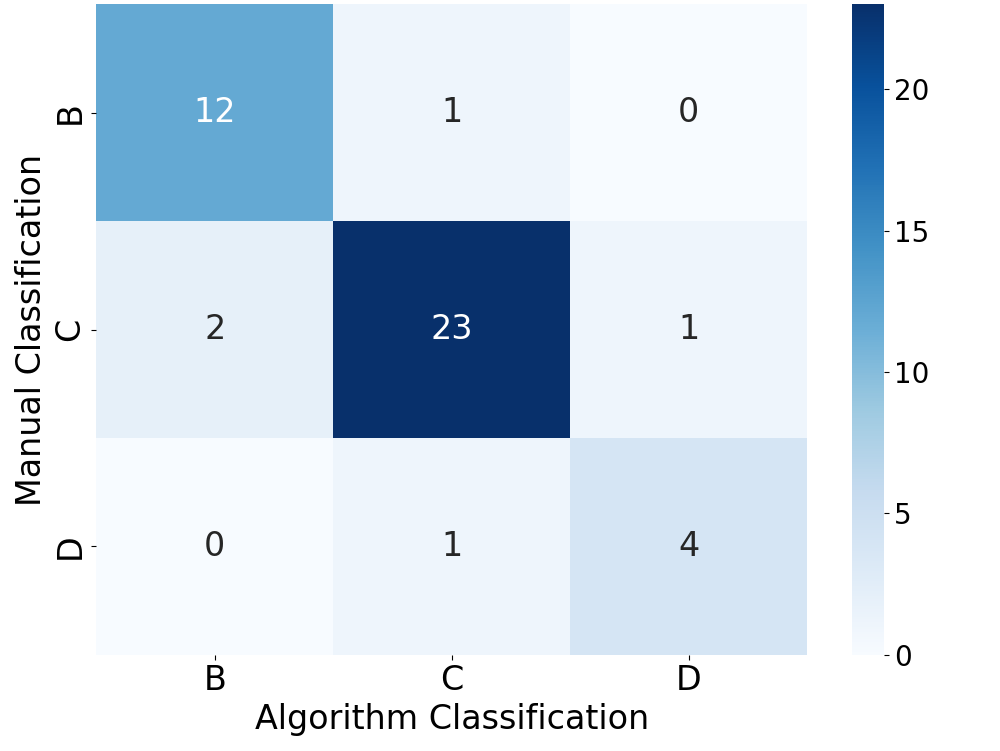} 
\caption{Confusion Matrix}  
\label{fig:confusion}  
\end{figure}

It can be observed from the confusion matrix (Fig \ref{fig:confusion}) that there’s no  misclassification between Class B and Class D, indicating that the algorithm can accurately distinguish between these two pose states, which have morphological differences. Specifically, the precision and recall for Class B are 86\% and 92\%, respectively, with an F1 score of 0.89, showing the algorithm’s strong capability in recognizing this class, with only a small number of misclassifications. Class D, on the other hand, exhibits balanced classification performance, with both precision and recall at 80\%, and an F1 score of 0.80.

The classification performance of Class C is the most outstanding, with the algorithm achieving a precision of 92\% and a recall of 88\%, resulting in an F1 score of 0.90. Considering that this state occurs most frequently during robot operations and requires timely identification and correction, this excellent classification accuracy is of significant importance for the practical application of the system.

In summary, by comparing with the manual classification results, the experimental results validate the effectiveness of the algorithm in the interventional robot pose state perception task. The algorithm not only performs excellently in terms of overall accuracy, but also demonstrates stable performance across all categories, particularly excelling in distinguishing the most common pose states. This fully demonstrates its potential for practical applications. These results provide reliable technical support for the development of subsequent robot control systems.

%%%%%%%%%%%%%%%%%%%%%%%%%%%%%%%%%%%%%%%%%%%%%%%%%%%%%%%%%%%%%%%%%%%%%%%%%%%%%%%%
\section{DISCUSSION \& CONCLUSION}
The method proposed in this study tackles the challenging problem of accurate trajectory tracking and pose state perception of the interventional robot. By performing precise image segmentation and extracting its skeleton structure, the presented algorithm effectively simplify the image information and carefully process the structural details to ensure the accuracy of the calculations. Furthermore, by respectively fitting lines of the robot's head and its surrounding vascular regions, the method obtains the key pose parameters, calculates the angles precisely, and classifies the pose state to support the subsequent control.

\begin{figure}[htbp]
\centering 
\includegraphics[width=0.25\textwidth]{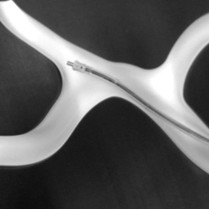} 
\caption{Situation when just entering a branch}
\label{fig:enter_fork} 
\end{figure}

The experimental results indicate that, compared to manual measurements and classifications, the algorithm shows no significant difference. Considering the potential for random errors and visual biases that can occur during manual recognition, the algorithm’s performance is, in fact, superior to manual operation. For instance, when just entering a branch (Fig \ref{fig:enter_fork}), human might be distracted by large empty areas, leading to decision-making errors. In contrast, the algorithm can maintain precise calculation of current path, unaffected by external noise. Further analysis reveals that the relatively large deviation in the angle perception experiment is mainly due to this kind of difference in strategies. However, it is worth noticing that this doesn’t imply a fault on either side, but rather a difference in approach. Overall, the experimental results provide strong evidence for the effectiveness of the proposed method, which perceives pose states through fitting, and shows its exceptionally high precision.

Meanwhile, the proposed pose state classification provides a foundation for developing control strategies, including the adjustment of the robot’s forward speed and steering speed, with the goal of achieving rapid movement while avoiding collision with the vascular walls. Although the current work does not include the actual design of the control module, the proposed image processing algorithm already meets the requirements for integration into a control system. Future research will focus on combining this algorithm with control theory, further developing and optimizing the control module, and ultimately implementing a closed-loop control system for the robot, enabling it to autonomously adjust in real-time within complex vascular environments.

%%%%%%%%%%%%%%%%%%%%%%%%%%%%%%%%%%%%%%%%%%%%%%%%%%%%%%%%%%%%%%%%%%%%%%%%%%%%%%%%

\end{document}